\documentclass[runningheads]{llncs}

\usepackage{eccv}

\usepackage{eccvabbrv}

\usepackage{graphicx}
\usepackage{booktabs}

\usepackage[accsupp]{axessibility}  

\usepackage{amsmath,amssymb}
\usepackage{mathtools}
\usepackage[ruled,vlined]{algorithm2e}
\usepackage{xcolor}
\usepackage{color}
\usepackage{tabularray}
\usepackage{bm}
\usepackage{multirow}
\usepackage{multicol}
\usepackage{float}
\usepackage{wrapfig}

\usepackage{amssymb}
\usepackage{amsmath,amsfonts}
\usepackage{amsopn}
\usepackage{bm} 
\usepackage{multirow}
\usepackage{flushend}
\usepackage{tabularx}

\usepackage{lipsum}

\newcommand{\vct}[1]{\boldsymbol{#1}} 

\newcommand{\ProbOpr}[1]{\mathbb{#1}}

\newcommand{\expect}[2]{%
\ifthenelse{\equal{#2}{}}{\ProbOpr{E}_{#1}}
{\ifthenelse{\equal{#1}{}}{\ProbOpr{E}\left[#2\right]}{\ProbOpr{E}_{#1}\left[#2\right]}}} 

\newcommand{\vmu}{\vct{\mu}}

\newcommand{\vp}{\vct{p}}
\newcommand{\vm}{\vct{m}}

\newcommand{\vx}{{\vct{x}}}

\newcommand{\vz}{{\vct{z}}}
\newcommand{\eat}[1]{}

\renewcommand{\paragraph}[1]{\vspace{.5em}\noindent\textbf{#1.}}

\usepackage[pagebackref,breaklinks,colorlinks,citecolor=eccvblue,linkcolor=Blue]{hyperref}       %

\usepackage{orcidlink}
\usepackage[stable]{footmisc}

\begin{document}

\title{Structural Assessment for\\ Understanding and Guiding Dataset Distillation in Discrete Token Space
}

\titlerunning{Structural Assessment for Dataset Distillation}

\author{
Yue Cao\inst{1} \and Jianyang Gu\inst{2} \and Vyacheslav Kungurtsev\inst{3} \and Yu Hu\inst{1} \and\\ Jozsef Hamari\inst{4} \and Zheng Liu\inst{1}\thanks{Corresponding authors.} \and Mohsen Zardadi\inst{4}\footnotemark[1]}

\authorrunning{Y.~Cao et al.}

\institute{The University of British Columbia \and
The Ohio State University \and
Czech Technical University in Prague \and TerraSense Analytics \\
\email{\{yue.cao, zheng.liu\}@ubc.ca, mohsen.zardadi@terrasense.ca}
}

\maketitle

\begin{abstract}
Dataset distillation (DD) has proven to reduce training cost while preserving accuracy. While promising, the factors that make one distilled dataset more effective than another remain poorly understood. In this work, we investigate this question through the lens of discrete visual tokenizers. Whereas many prior DD efforts emphasize matching global data distributions, we suggest that the effectiveness depends on which semantic concepts are captured and how they are composed. Discrete visual tokenizers provide a finite vocabulary that enables direct statistical analysis of such compositional structure. Through quantitative analysis of token-level statistics, we introduce the \textbf{structural score} to measure the adequacy of token compositions. We observe that distilled datasets with balanced token composition yield higher validation performance. On the other hand, divergence from the original data does not necessarily harm performance. We further show that samples with high structural scores in the discrete token space can effectively guide diffusion-based DD. Our findings highlight the importance of token composition in dataset effectiveness, offering a principled complement to distributional similarity considerations in DD.

\end{abstract}
\section{Introduction}
\label{sec:intro}


    
Dataset distillation aims to replace a large training set with a surrogate while preserving training performance~\cite{DD_Review1,DD_Review2,Rebuttal_New_Overview}. Despite rapid empirical progress, what makes a \textit{good} distilled dataset remains poorly understood~\cite{lu2023can,lu2025dataset}. Many prior efforts enforce surrogates to match the continuous distributions of real data via feature matching \cite{DM_Zhao,Dream-Kmeans-VAE,FRePo_Zhou}, optimal transport \cite{Wasserstein_Liu}, or gradient alignment \cite{Gradient_Matching_Zhao,Matching_Trajectories_Cazenavette}. However, such distributional proximity alone does not indicate semantic structures or discriminate variations informative for learning. 
A surrogate may be close to real data in embedding space yet still collapse rare concepts, overrepresent trivial structures, or fail to span discriminative variations. 
In practice, the dataset most similar to the original is not always the most effective for training. This observation raises an open question: what properties of a surrogate, beyond distributional similarity, ultimately govern its training effectiveness?
\begin{figure}[t]
    \centering
    \includegraphics[width = 1.0\linewidth]{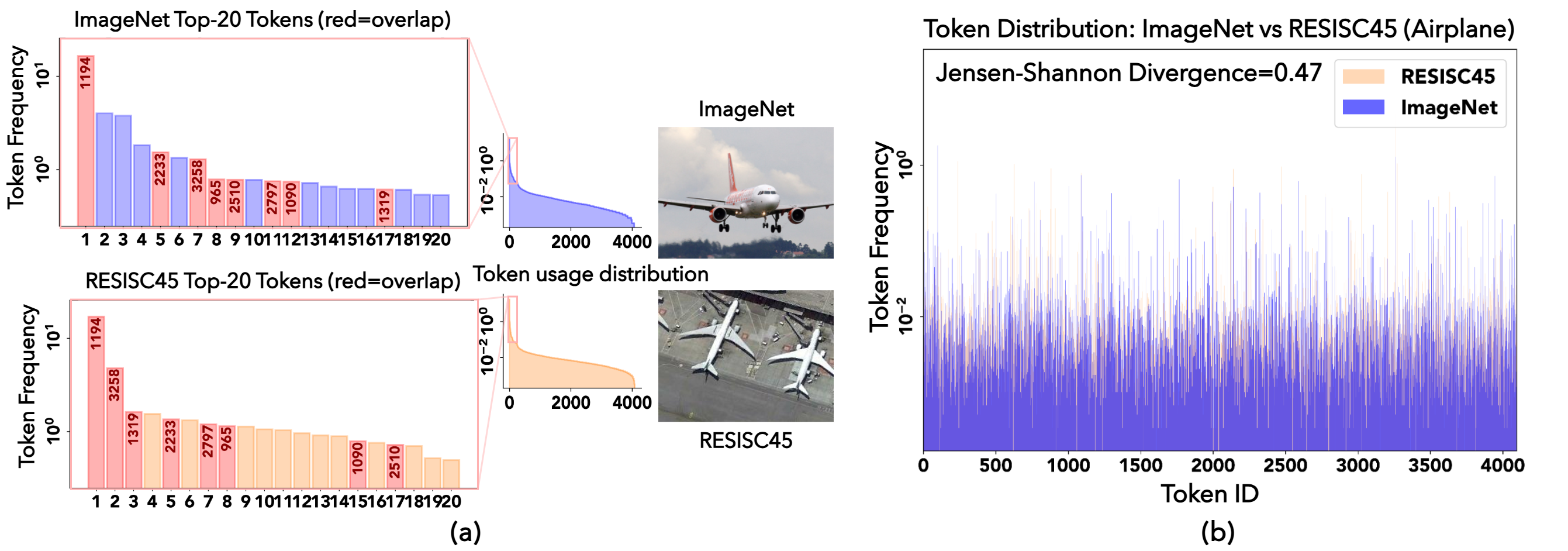}
    \caption{\textbf{The discrete token distribution of airplane images in standard and remote sensing imagery}. (a) The airplanes frequently use similar top tokens across datasets. (b) At the same time, they exhibit a large divergence (JSD=0.47) in their overall token distribution due to structural differences.}
    \label{fig:domain-diff}
    \vspace{-12pt}
\end{figure}

We investigate this question through the lens of \textbf{discrete visual tokenizers}. Unlike high-dimensional continuous feature spaces, where semantics are heavily entangled, discrete tokenizers map each image into a sequence of indices from a finite vocabulary. This enables direct statistical analysis of how visual primitives are composed within a dataset. As illustrated in \autoref{fig:domain-diff}, we use VQ-VAE~\cite{VAR_Tian,VQVAE_original} to quantize airplane images from ImageNet~\cite{ImageNet_Jia} and remote sensing~\cite{RESISC45_cheng} into visual tokens. Although these two domains differ substantially in viewpoint, background, and style, they share a subset of high-frequency tokens that likely correspond to common airplane-related visual primitives. At the same time, their overall token distributions diverge due to the differences in context and domain-specific artifacts. This motivates us to move beyond matching global distributions and instead explicitly characterize token structure within a dataset.


To formalize this assessment, we represent each image as a discrete distribution over the shared visual vocabulary and quantify this representation through three complementary metrics:
(1) Jensen--Shannon Divergence (JSD) \cite{JSD_Lin} between an image's token distribution and the average distribution of its corresponding class. Lower JSD indicates samples with more representative compositional patterns.
(2) Herfindahl--Hirschman Index (HHI) \cite{HHI_Hirschman} over the token distribution, where a lower HHI reflects more balanced usage of codebook tokens.
(3) Coverage rate (COV) over class-discriminative tokens identified from the real dataset, where higher coverage reflects stronger preservation of class-relevant primitives.
Together, these metrics provide a holistic structural signature of each image. We then relate the aggregated statistics of surrogates generated by diverse DD methods to their validation accuracy via regression analysis. 
Empirically, the combination of these three metrics, named the \textbf{structural score}, provides a solid prediction of validation accuracy.
The resulting coefficients reveal that more balanced token composition primarily leads to higher accuracy. Intriguingly, merely minimizing token-level divergence from the original set does not reliably correlate with better performance. 


Building on these insights, we move from diagnosis to design and explore how token-level statistics can guide surrogate generation. Unlike prior methods that rely on continuous feature centroids to guide diffusion~\cite{MGD3,D4M-Kmeans-VAE}, we cluster data in the discrete token space and rank samples based on the structural score to acquire optimal guidance signals. 
Using the same steering technique as prior methods, the \textbf{Token-Guided Dataset Distillation (TGDD)} method demonstrates more effective information integration across diverse diffusion backbones and benchmarks. On ImageWoof, TGDD yields a 5.4\% improvement compared to the state-of-the-art baseline~\cite{MGD3}.


To summarize, our work presents two contributions: (1) We introduce a training-free structural score that reliably assesses the quality of the distilled dataset in various DD methods and domains. (2) We develop a token-guided distillation framework that leverages this score to actively guide surrogate generation, proving the effectiveness of the discrete structural assessment.

\section{Related Work}
\label{sec:RelatedWork}

\subsection{Dataset Distillation}
Dataset distillation aims to synthesize a surrogate dataset such that training a model on it yields comparable performance to training on the full original dataset~\cite{DD_Review1,DD_Review2}.
Among previous efforts, optimization-based methods directly update surrogate images to mimic the training dynamics of the original dataset.
\cite{Gradient_Matching_Zhao, Matching_Trajectories_Cazenavette, IDC_Kim, vahidian2025group} align the gradient or training trajectory, and \cite{Wasserstein_Liu, DANCE_Zhang, RDED_Sun, Distribution_NCFM_Wang, OPTICAL_Cui,SReL_Yin} align feature or distribution statistics.
As optimization-based methods typically rely on a teacher model for supervision, their generalization across architectures is constrained. 
Another line of work synthesizes datasets using generative priors. For example, GLaD \cite{GLaD_generative_prior_Cazenavette}, D2M \cite{D2M_Sajedi}, and H-PD \cite{H-PD_Zhong} leverage pretrained GANs to facilitate the optimization of the dataset.
More recent works employ diffusion and visual autoregressive models as the backbone for DD, formulating synthesis as a guided generation process~\cite{Minimax_diffusion_gu,D4M-Kmeans-VAE,zhao2025taming,gu2025concord,zhao2026hieramp}.
MGD$^3$ \cite{MGD3} and VLCP \cite{vlcp_zou} improve intra-class diversity through mode guidance and vision–language prototypes.
IGD \cite{Influence_guided_diffusion_Chen} and Ca$\text{O}_{2}$ \cite{wang2025cao} further introduce trajectory-influence and consistency-based guidance.
Although representativeness, informativeness, and diversity are widely acknowledged to be important, prior work often relies on heuristic similarity scores as proxies. In this work, we aim to interpret datasets as compositions of visual concepts in certain contexts. Thereby, we turn the assessment of distilled data from distributional similarity to the structural score. 

\subsection{Feature-based Data Selection Criteria}
To assess the usefulness of features and data samples, prior work employs a range of statistical and information-theoretic measures. The Herfindahl Hirschman Index (HHI) \cite{HHI_Hirschman}, Shannon entropy \cite{shannon_entropy}, and the Gini coefficient \cite{gini_coefficient} are used to characterize the concentration and diversity of feature distributions. To compare probability distributions induced by features or by selected subsets of data, Jensen--Shannon divergence and Kullback--Leibler divergence are widely used \cite{KL_divergence, JSD_Lin}. For feature specificity, term frequency--inverse document frequency \cite{TF-IDF_Method_Sammut} highlights rare yet discriminative patterns. Collectively, these metrics provide quantitative tools for evaluating informativeness, coverage, and redundancy, which motivate the token-level criteria used in our framework.
\section{Structural Assessment for Dataset Distillation
}
\label{sec:Token-Guided-Assessment}

As motivated in \S\ref{sec:intro}, beyond overall proximity to the original data, we seek a more fine-grained understanding of surrogate quality. To this end, we propose explicitly assessing the structure of visual primitives by mapping continuous images into a discrete token space. Statistics over a finite vocabulary enable analyzing how these tokens are used at the dataset level. Building on this framework, we introduce the \textbf{structural score} as a robust signature of dataset quality to quantitatively evaluate the distilled surrogate.

\subsection{
Structural Representation in Token Space
}\label{sec:TokenStatistics}

We first employ a discrete visual tokenizer to map continuous images into a sequence of indices from a finite token vocabulary. In recent years, there have been extensive efforts in developing effective visual tokenizers~\cite{VAR_Tian,VQGAN_Esser,BeitV2_Peng}. This tokenizing step does not rely on a specific tokenizer type, and can be broadly applied. Without loss of generality, we first demonstrate an instantiation of the assessment with a multi-scale VQ-VAE~\cite{VAR_Tian} model.

Given a VQ-VAE codebook of size $V$ and the encoder with $L$ scales, each surrogate image is embedded with the encoder into discrete token maps $\{\vz_i^{(1)}, \vz_i^{(2)}, \allowbreak \ldots,  \vz_i^{(L)}\}$, where $\vz_i^{(\ell)}$ contains indices from the codebook at the $\ell$-th scale of the total $L$ scales. 
For each scale, token occurrences are counted and normalized to form a probability vector over the codebook:
\[
\vp_i^{(\ell)} \in \mathbb{R}^{V}, 
\qquad 
\sum_{k=1}^{V} p_i^{(\ell)}(k)=1,
\]
where $p_i^{(\ell)}(k)$ denotes the relative frequency of the $k$-th token in image $\vx_i$ at scale $\ell$.
While VQ-VAE tokens on different scales capture distinct visual information, analyzing them separately would yield fragmented scores that are difficult to reconcile. 
To synthesize these signals into a coherent statistical signature, we aggregate the distributions across all scales. Since the number of tokens grows quadratically with scale, a direct summation causes an inherent density imbalance.
Therefore, we fuse the token distributions using a weighted sum:
\[
\vp_i = \sum_{\ell=1}^{L} w_\ell \, \vp_i^{(\ell)}, 
\]
where $\vp_i$ represents the fused token prior. 
We group the $L = 10$ scales into low (1--3), mid (4--7), and high (8--10) resolution bands. 
Intuitively, we assign them relative weights of $[3, 1, 0.5]$. 
This strategy effectively upweights the low-resolution bands to compensate for their sparsity.
The validation of this weighting scheme is provided in Appendix \S\ref{sec:supp-ScaleWeightingConfig}.

Similar to words, each token in the codebook may convey a series of semantics. The semantics are not directly interpretable without being incorporated into a composition. 
We then examine the utilization structure of different visual tokens in addition to their presence. 
Specifically, we look into three properties of the distilled dataset that characterize its token-level composition:

\paragraph{Contextual fit} The surrogate dataset is expected to use a composition of tokens similar to the original dataset to reflect the corresponding contexts. Jensen--Shannon Divergence (JSD) \cite{JSD_Lin} is incorporated to measure the contextual fit:
\[
\mathrm{JSD}(\vp_i, \vmu_c) = \tfrac{1}{2}\left(\mathrm{KL}(\vp_i \,\|\, \vm) + \mathrm{KL}(\boldsymbol{\mu}_c \,\|\, \vm)\right),
\]
where $c$ is the class that $\vx_i$ belongs to, and $\vmu_c$ is the average token distribution of the corresponding class, \ie, the class centroid. $\mathrm{KL}(\cdot\,\|\,\cdot)$ calculates the KL divergence between two distributions, and $\vm$ is the mixture distribution of $\vp_i$ and $\vmu_c$. Lower JSD scores indicate that the samples have more typical compositional patterns of that class.

\paragraph{Compositional richness} The surrogate should convey meanings by combining a broad set of distinct primitives rather than reusing a few. Herfindahl-Hirschman Index (HHI) \cite{HHI_Hirschman} is employed for this measurement:
\[
\mathrm{HHI}(\vp_i)
= \sum_{k=1}^{V} \big(p_i(k)\big)^2,
\]
where $k\in\{1,\dots,V\}$ indexes codebook tokens, and $p_i(k)$ is the $k$-th entry of the fused per-image token distribution $\vp_i$. Smaller HHI values correspond to more balanced token usage and thus richer visual information within the image.

\paragraph{Categorical presence} The usage of tokens should carry important class characteristics. We use the coverage rate of class-related tokens identified by TF-IDF \cite{TF-IDF_Method_Sammut} that appear in the original dataset:
\[
\mathrm{COV}(\vp_i)
= \sum_{k \in \mathcal{T}_c} p_i(k),
\]
where $\mathcal{T}_c \subseteq \{1,\dots,V\}$ denotes the set of top TF-IDF tokens for class $c$. Higher coverage indicates that the sample uses more class-relevant visual primitives.

These three metrics reflect complementary properties of datasets.
Note that the structural assessment can also be implemented with other discrete visual tokenizers, such as VQGAN~\cite{VQGAN_Esser} and BEiTv2~\cite{BeitV2_Peng}. We then investigate the impact of these three properties on the validation performance.

\subsection{Empirical Results}
\label{sec:EmpiricalResults}

We collect surrogates for ImageWoof generated by a variety of dataset distillation methods, including both optimization-based (DM~\cite{DM_Zhao}, SRe$^2$L~\cite{SReL_Yin}), selection-based (RDED~\cite{RDED_Sun}), and generative-based approaches (MGD$^3$~\cite{MGD3}, Minimax~\cite{Minimax_diffusion_gu}, VAR~\cite{VAR_Tian}, Stable Diffusion~\cite{stable_diffusion_3_5}). For each method except SRe$^2$L, we collect multiple surrogates from at least 5 runs to reduce randomness. 
We compute the token statistics of each surrogate with respect to the original ImageWoof \cite{imageWoofNette} training set, where each metric is computed by averaging the scores of all individual samples within the distilled dataset. We also train a ResNet-10 \cite{resnet_he} classifier from scratch on each surrogate to obtain their validation accuracy. 


\begin{wrapfigure}{R}{0.55\textwidth}
  \centering
  \vspace{-20pt}
   \includegraphics[width=0.9\linewidth]{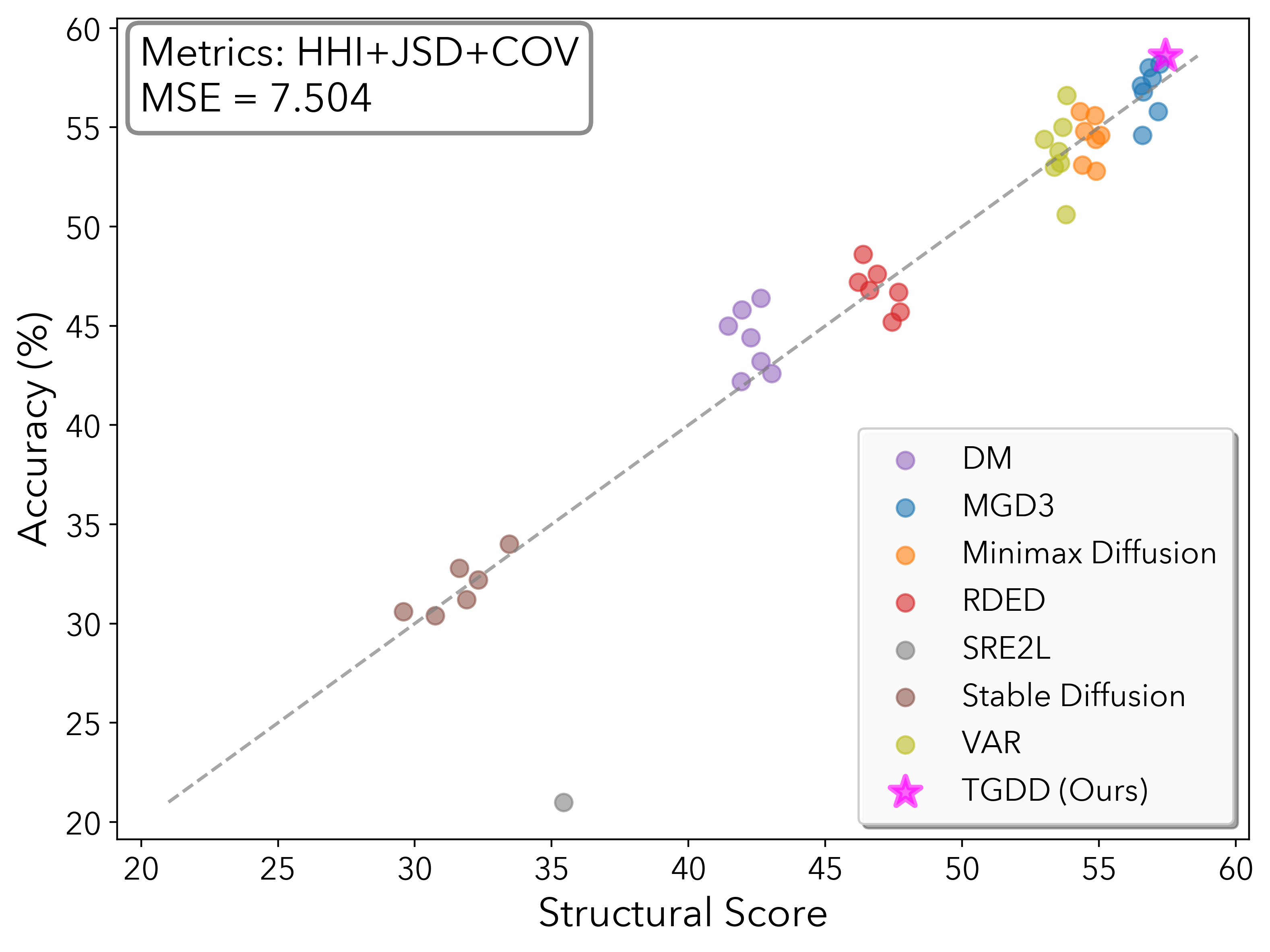}
  \vspace{-8pt}
   \caption{Structural score versus ground-truth validation accuracy on ImageWoof across distilled datasets generated by multiple methods. We report the mean-squared error (MSE) over all data points. Note that TGDD (Ours) is only used for test but not for linear regression. }
   \label{fig:DatasetQualityScore}
  \vspace{-18pt}
\end{wrapfigure}

We fit a linear regression model with a Lasso regularization coefficient of $0.5$ that maps the token statistics to the resulting validation accuracy.
As shown in \autoref{fig:DatasetQualityScore}, the predicted validation accuracy exhibits a small mean-squared error (MSE) from the ground-truth accuracy. 
This strong correlation indicates that the token statistics provide essential information for assessing the quality of surrogates.

\begin{figure*}[t]
  \centering
\vspace{-8pt}
   \includegraphics[width=1.0\linewidth]{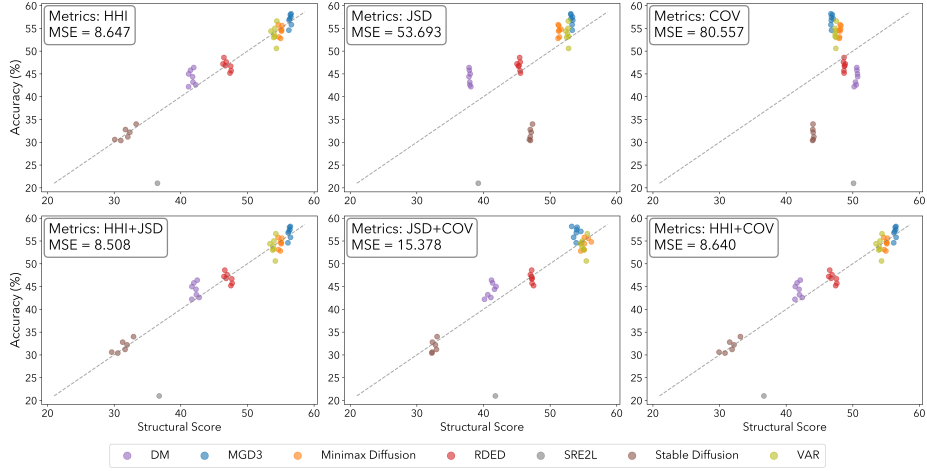}
\vspace{-16pt}
   \caption{The ablation study of different combinations of token statistics (HHI, JSD, COV) in predicting model accuracy via linear regression. Each subplot shows the relationship between the adopted token statistics and model accuracy across various distilled datasets, with Mean Squared Error (MSE) reported. }
   \label{fig:assessment_abl}
\vspace{-6pt}
\end{figure*}

We further ablate this linear regression with different groups of metrics, and the results are shown in \autoref{fig:assessment_abl}.
When only one metric is adopted to fit the validation accuracy, HHI illustrates a smaller MSE compared with JSD and COV. 
Combining two metrics yields a smaller MSE compared with each of them alone. The best explanation is produced from the combination of all three metrics, which is our proposed assessment in \autoref{fig:DatasetQualityScore}. The acquired coefficients for each metric and the intercept $\beta$ are:
\[
w_{\mathrm{JSD}}=2.96,\\ w_{\mathrm{HHI}}=-10.99,\\ w_{\mathrm{COV}}=2.07,\\ \beta=47.75.
\]
While they cannot be evaluated in isolation, the validation performance tends to have a strong correlation with lower HHI values, \ie, more balanced token compositions. We refer to this combination of statistics as the \textbf{structural score}, which provides more comprehensive assessments of distilled datasets compared with the distributional similarity alone. 

\begin{wraptable}{R}{0.6\textwidth}
\vspace{-16pt}
    \caption{Detailed metric values and validation accuracy of 50-IPC surrogates for ImageWoof.}
    \label{tab:case-study}
    \centering
    \small
\setlength{\tabcolsep}{4pt}
    \resizebox{\linewidth}{!}{
    \begin{tabular}{l|cccc}
    \toprule
        Method & JSD & HHI & COV & Accuracy \\
    \midrule
        Stable Diffusion~\cite{stable_diffusion_3_5} & $0.122$ & $4.608\times10^{-3}$ & $0.299$ & $31.2$ \\
        DM~\cite{DM_Zhao} & $0.186$ & $3.334\times10^{-3}$ & $0.207$ & $42.2$ \\
        RDED~\cite{RDED_Sun} & $0.136$ & $2.616\times10^{-3}$ & $0.228$ & $47.2$ \\
        MGD$^3$~\cite{MGD3} & $0.077$ & $1.185\times10^{-3}$ & $0.261$ & $55.8$ \\
    \bottomrule
    \end{tabular}
    }
\vspace{-10pt}
\end{wraptable}

We list the token statistics and validation accuracy of surrogates distilled by Stable Diffusion \cite{stable_diffusion_3_5}, DM \cite{DM_Zhao}, RDED \cite{RDED_Sun}, and MGD$^3$ \cite{MGD3} in \autoref{tab:case-study} for a more detailed case study. Aligned with the coefficient, HHI has the strongest influence on the validation accuracy, while the other two metrics cannot lead to a strong correlation by themselves. 
Notably, DM and RDED illustrate higher JSD compared with the images generated by Stable Diffusion, but still yield higher accuracy. 
This result corresponds to our claim that the distributional similarity alone cannot indicate the data effectiveness. 


\paragraph{Generalizability of the scoring model}
With the initial metric coefficients acquired on ImageWoof, we further conduct the same linear regression with surrogates from ImageNette~\cite{imageWoofNette}. The obtained weights for the three metrics are: $w_{\mathrm{JSD}}=2.43,\; w_{\mathrm{HHI}}=-9.59,\; w_{\mathrm{COV}}=1.41.$
While the absolute numbers vary, the relative importance of the metrics remains consistent, with HHI suggesting the most significant influence on the validation performance. This result confirms the effectiveness of the derived conclusions from the structural assessments.

\begin{table}[h]
\centering
\caption{Application of the proposed structural score to the EuroSAT dataset, demonstrating that our metric generalizes to visually distinct domains without retraining.}
\label{tab:eurosat}

\setlength{\tabcolsep}{6pt}
\small
\resizebox{\textwidth}{!}{
\begin{tabular}{c|cccccccccc}
\toprule
Iters & 0 & 200 & 400 & 600 & 800 & 1000 & 1200 & 1400 & 1600 & 1800 \\
\midrule
Structural Score & -76.7 & -34.2 & -22.1 & -15.5 & -13.7 & -11.1 & -9.9 & -4.9 & -1.6 & 2.2 \\
Accuracy (\%) & 11.5 & 40.6 & 44.9 & 47.1 & 47.8 & 51.7 & 53.0 & 55.8 & 57.2 & 61.1 \\
\bottomrule
\end{tabular}
}
\end{table}

\paragraph{Direct application to other domains}
Although the relative relationship stands across datasets, it is infeasible to calculate the coefficients for each new application. Therefore, we demonstrate the direct generalization of the acquired coefficients on remote sensing imagery~\cite{Eurosat_Helber}. 
We employ Distribution Matching (DM), an optimization-based method, to perform the dataset distillation for EuroSAT. During the distillation process, we collect 10 surrogate datasets at different timesteps. The structural score and corresponding validation accuracy of each surrogate are reported in  \autoref{tab:eurosat}. As the images are progressively updated, the structural scores increase monotonically alongside the downstream validation accuracy.
Due to the large domain gap, the predicted score does not equal the accuracy. However, it is still able to compare the effectiveness of different surrogates based on relative relationships. 
This trend confirms that our token-level metrics are robust indicators of dataset quality, even when applied to domains significantly different from ImageNet.

\begin{figure*}[t]
  \centering
   \includegraphics[width=1.0\linewidth]{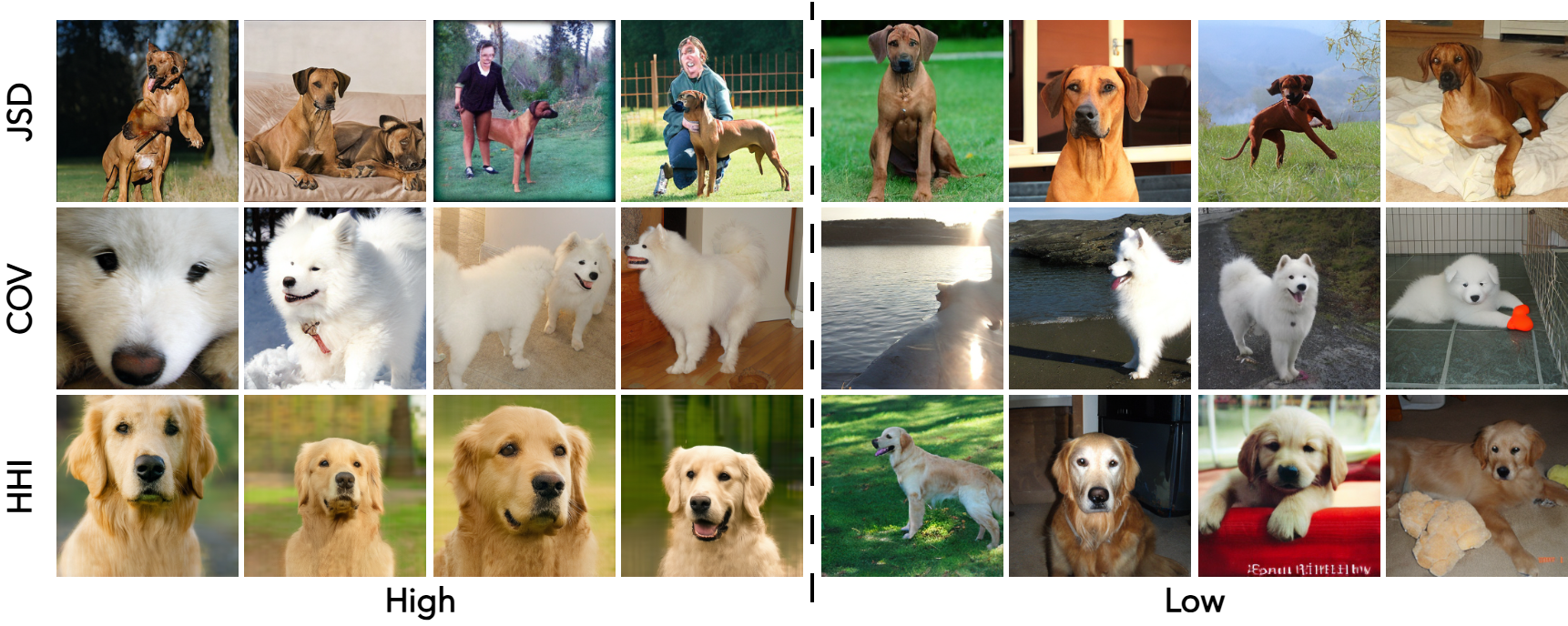}
\vspace{-16pt}
   \caption{Qualitative comparison of images with high and low values of JSD, COV, and HHI metrics. Each row corresponds to one metric, with images on the left exhibiting high values and those on the right showing low values.}
   \label{fig:Example_Vis}
\vspace{-6pt}
\end{figure*}

\subsection{Example Visualization}

We compare representative image groups with high and low metric values to better understand how each token statistics in the structural assessment reflects dataset quality, as illustrated in \autoref{fig:Example_Vis}. We select examples generated by MGD$^3$ \cite{MGD3} and Stable Diffusion \cite{stable_diffusion_3_5}, and analyze how each metric correlates with visual quality. For JSD, both sets of images are generated by MGD$^3$, and the JSD is computed by comparing each image's token distribution with the average token distribution of the corresponding class in the original ImageWoof dataset. Images with higher JSD values tend to exhibit unrealistic features, such as disordered object shapes or incorrect combinations of semantic elements, making them deviate from the original class distribution. For the coverage metric, lower values are often associated with images where the target object occupies a smaller region of the image or is missing entirely. This indicates a weaker representation of class-specific content. Finally, we visualize images based on their HHI values. We use samples generated by MGD$^3$ for the low-HHI group, which generally exhibit higher diversity. By contrast, samples generated by Stable Diffusion show similar object poses and more repetitive patterns, reflected by higher HHI values.

\section{Token-Guided Dataset Distillation}
\label{sec:Method}

The preceding analysis demonstrates that token statistics reliably indicate the quality of distilled datasets. Building on this insight, we further show that these statistics can also be incorporated to guide diffusion denoising.
Following~\cite{MGD3}, we cluster each class to identify representative modes, but based on the token distribution instead of continuous features. 
We then rank and select anchors from each cluster using the proposed structural score, and use the selected anchors to guide a diffusion model in generating synthetic samples as in~\cite{MGD3}. 

\subsection{Anchor Selection}
\label{sec:TokenDM}

Given a real dataset $\mathcal{T} = \{(x_i, y_i)\}_{i=1}^{N_T}$, the discrete visual tokenizer is employed to embed each image to capture the underlying distribution of visual concepts $\{\vp_i\}$.
For numerical stability, principal component analysis (PCA) is first applied to reduce the token space to $d$ dimensions, and the projected vectors are L2-normalized. 
This normalization makes Euclidean distances a monotonic transformation of cosine similarities, ensuring that clustering captures differences in token composition rather than in feature magnitude.
$k$-means is then run on the normalized vectors to obtain $K_c$ clusters, where $K_c$ is set by the IPC target.
This step is the same as that in~\cite{MGD3}, except in the discrete token space, where distances reflect composition differences over visual concepts. Therefore, each cluster corresponds to a pattern of token use within the class.


Token-level clustering reveals intra-class token modes but leaves many candidates per mode. 
Instead of using the centroid as in~\cite{MGD3}, we select a small number of anchors from each cluster based on the token statistics identified in \S\ref{sec:TokenStatistics}. Specifically, we use the linear regression model in~\S\ref{sec:EmpiricalResults} to predict a structure score for each sample. The only difference from the previous JSD calculation is that samples are compared with cluster centers instead of class centers. 
The top $M$ images with the highest scores are selected as anchors. We set $M=20$ anchors per cluster for IPC $<$ 50 and $M=10$ for higher IPC settings.
We summarize the anchor selection procedure in \autoref{alg:anchor_selection} of Appendix \S\ref{sec:supp-algorithm}.

\subsection{Token-Guided Generation}

The selected anchors from each cluster are used to guide synthetic data generation. 
We average the anchor embeddings of one cluster into one latent, representing the corresponding mode.
Following the denoising procedure of \cite{MGD3}, we guide the denoising process with the acquired modes:
\[
\vz_{t-1} = \vz_{t-1}^{\text{base}} + \mathbb{1}_{\{t>t_{\text{stop}}\}}\,\lambda\,\big(\bar{\vz}_{c,m} - \hat{\vz}_{0}\big),
\]
where $\lambda$ denotes the mode-guidance strength, $\vz_t$ is the latent variable at step $t$, and $\hat{\vz}_{0}$ is the predicted clean latent. The guidance is applied only for timesteps $t > t_{\text{stop}}$ to maintain sample diversity. 
We demonstrate that by simply summarizing modes in the discrete token space with the help of token statistics, the acquired guidance can be more effective than that of continuous cluster centroids. 
Moreover, in \S\ref{sec:ablation}, we demonstrate that the method can be applied to a broad range of discrete tokenizers and diffusion architectures to provide more effective guidance than continuous models.

\section{Experiments}
\label{sec:Experiments}

\subsection{Implementation Details}
For the visual tokenizer, we adopt the multi-scale VQ-VAE model developed by \cite{VAR_Tian}, while the diffusion generation framework employs a pre-trained DiT \cite{DiT_Peebles}. Following the literature \cite{MGD3,DiT_Peebles}, we use 256×256 image resolution with 50 sampling steps and apply stop guidance at step 25. All experiments are conducted on a single RTX 3080 GPU. Further implementation details are provided in Appendix~\S\ref{sec:supp-ImplementationDetails}.

\paragraph{Datasets}
We evaluate our method on multiple high-resolution benchmarks with 256×256 images to thoroughly validate its performance. The dataset used in this experiment includes ImageNette, ImageWoof \cite{imageWoofNette}, ImageIDC \cite{IDC_Kim}, ImageNet-100 \cite{imagenet-100}, and ImageNet-1k \cite{ImageNet_Jia}. ImageNette and ImageIDC each contain 10 classes, and ImageWoof comprises 10 different dog breeds, making it particularly challenging due to high inter-class similarity. 
For evaluation, we adopt the same metrics as \cite{Minimax_diffusion_gu,MGD3}.

\begin{table*}[t]
\centering
\caption{Comparison of performance across state-of-the-art methods on ImageWoof. The best results are marked in \textbf{bold}.}
\label{tab:imagewoof}
\small
\vspace{-6pt}
\setlength{\tabcolsep}{4pt}
\resizebox{\textwidth}{!}{%
\begin{tabular}{@{}ll|ccccccc|c@{}}
\toprule
IPC (Ratio) & Test Model & Random & Herding \cite{Herding_welling} & DiT \cite{DiT_Peebles} & DM \cite{DM_Zhao}  & MiniMax \cite{Minimax_diffusion_gu} & MGD$^3$ \cite{MGD3} & \textbf{TGDD} & Full \\
\midrule
\multirow{3}{*}{10 (0.8\%)}  & ConvNet-6   & $24.3_{\pm 1.1}$ & $26.7_{\pm 0.5}$ & $34.2_{\pm 1.1}$      & $26.9_{\pm 1.2}$              & $\mathbf{37.0}_{\pm 1.0}$   & $34.7_{\pm 1.1}$       & $34.8_{\pm 1.1}$             & $86.4_{\pm 0.2}$ \\
            & ResNetAP-10 & $29.4_{\pm 0.8}$ & $32.0_{\pm 0.3}$ & $34.7_{\pm 0.5}$      & $30.3_{\pm 1.2}$             & $39.2_{\pm 1.3}$   & $40.4_{\pm 1.9}$       & $\mathbf{41.2}_{\pm 2.6}$    & $87.5_{\pm 0.5}$ \\
            & ResNet-18   & $27.7_{\pm 0.9}$ & $30.2_{\pm 1.2}$ & $34.7_{\pm 0.4}$      & $33.4_{\pm 0.7}$           & $37.6_{\pm 0.9}$   & $38.5_{\pm 2.5}$       & $\mathbf{38.8}_{\pm 0.9}$    & $89.3_{\pm 1.2}$ \\
\midrule
\multirow{3}{*}{20 (1.6\%)}  & ConvNet-6   & $29.1_{\pm 0.7}$ & $29.5_{\pm 0.3}$ & $36.1_{\pm 0.8}$      & $29.9_{\pm 1.0}$       & $37.6_{\pm 0.2}$   & $39.0_{\pm 3.5}$       & $\mathbf{39.1}_{\pm 0.6}$           & $86.4_{\pm 0.2}$ \\
            & ResNetAP-10 & $32.7_{\pm 0.4}$ & $34.9_{\pm 0.1}$ & $41.1_{\pm 0.8}$      & $35.2_{\pm 0.6}$                 & $45.8_{\pm 0.5}$   & $43.6_{\pm 1.6}$       & $\mathbf{46.3}_{\pm 0.8}$             & $87.5_{\pm 0.5}$ \\
            & ResNet-18   & $29.7_{\pm 0.5}$ & $32.2_{\pm 0.6}$ & $40.5_{\pm 0.5}$      & $29.8_{\pm 1.7}$       & $42.5_{\pm 0.6}$   & $41.9_{\pm 2.1}$       & $\mathbf{42.7}_{\pm 0.5}$    & $89.3_{\pm 1.2}$ \\
\midrule
\multirow{3}{*}{50 (3.8\%)}  & ConvNet-6   & $41.3_{\pm 0.6}$ & $40.3_{\pm 0.7}$ & $46.5_{\pm 0.8}$      & $44.4_{\pm 1.0}$    & $53.9_{\pm 0.6}$   & $54.5_{\pm 1.6}$       & $\mathbf{54.9}_{\pm 0.7}$    & $86.4_{\pm 0.2}$ \\
            & ResNetAP-10 & $47.2_{\pm 1.3}$ & $49.1_{\pm 0.7}$ & $49.3_{\pm 0.2}$      & $47.1_{\pm 1.1}$      & $56.3_{\pm 1.0}$   & $56.5_{\pm 1.9}$       & $\mathbf{60.3}_{\pm 0.9}$   & $87.5_{\pm 0.5}$ \\
            & ResNet-18   & $47.9_{\pm 1.8}$ & $48.3_{\pm 1.2}$ & $50.1_{\pm 0.5}$      & $46.2_{\pm 0.6}$     & $57.1_{\pm 0.6}$   & $58.3_{\pm 1.4}$       & $\mathbf{61.5}_{\pm 0.4}$    & $89.3_{\pm 1.2}$ \\
\midrule
\multirow{3}{*}{70 (5.4\%)}  & ConvNet-6   & $46.3_{\pm 0.6}$ & $46.2_{\pm 0.6}$ & $50.1_{\pm 1.2}$      & $47.5_{\pm 0.8}$    & $55.7_{\pm 0.9}$   & $55.1_{\pm 2.5}$       & $\mathbf{58.1}_{\pm 1.5}$   & $86.4_{\pm 0.2}$ \\
            & ResNetAP-10 & $50.8_{\pm 0.6}$ & $53.4_{\pm 1.4}$ & $53.4_{\pm 0.9}$      & $51.7_{\pm 0.8}$      & $58.3_{\pm 0.2}$   & $60.2_{\pm 2.4}$       & $\mathbf{63.4}_{\pm 1.0}$    & $87.5_{\pm 0.5}$ \\
            & ResNet-18   & $52.1_{\pm 1.0}$ & $49.7_{\pm 0.8}$ & $51.5_{\pm 1.0}$      & $51.9_{\pm 0.8}$      & $58.8_{\pm 0.7}$   & $59.7_{\pm 2.7}$       & $\mathbf{65.1}_{\pm 1.2}$    & $89.3_{\pm 1.2}$ \\
\midrule
\multirow{3}{*}{100 (7.7\%)} & ConvNet-6   & $52.2_{\pm 0.4}$ & $54.4_{\pm 1.1}$ & $53.4_{\pm 0.3}$      & $55.0_{\pm 1.3}$         & $61.1_{\pm 0.7}$   & $60.1_{\pm 1.2}$       & $\mathbf{63.6}_{\pm 1.6}$    & $86.4_{\pm 0.2}$ \\
            & ResNetAP-10 & $59.4_{\pm 1.0}$ & $61.7_{\pm 0.9}$ & $58.3_{\pm 0.8}$      & $56.4_{\pm 0.8}$      & $64.5_{\pm 0.2}$   & $66.5_{\pm 1.0}$       & $\mathbf{67.3}_{\pm 0.8}$    & $87.5_{\pm 0.5}$ \\
            & ResNet-18   & $61.5_{\pm 1.3}$ & $59.3_{\pm 0.7}$ & $58.9_{\pm 1.3}$      & $60.2_{\pm 1.0}$         & $65.7_{\pm 0.4}$   & $68.8_{\pm 0.7}$       & $\mathbf{70.1}_{\pm 0.6}$    & $89.3_{\pm 1.2}$ \\
\bottomrule
\end{tabular}
}
\vspace{-6pt}
\end{table*}

\subsection{Comparison with State-of-the-art Methods}

We compare our method against multiple state-of-the-art baselines using a consistent evaluation protocol. The baselines include both generative-based methods: Minimax Diffusion \cite{Minimax_diffusion_gu}, DiT \cite{DiT_Peebles}, and MGD$^3$ \cite{MGD3}, and optimization-based methods: DM \cite{DM_Zhao}, RDED \cite{RDED_Sun}, SRe$^2$L \cite{SReL_Yin} and IDC \cite{IDC_Kim}. All methods are evaluated across multiple images-per-class (IPC) settings and classification architectures to validate the generalizability.

ImageWoof is selected for initial analysis due to its fine-grained classification structure, where all classes correspond to different dog breeds. The high visual similarity among these classes requires models to rely on subtle, localized features for accurate discrimination. In such scenarios, methods that focus solely on global distribution alignment often struggle to capture meaningful intra-class variation, leading to trivial or overlapping representations. TGDD addresses this challenge by leveraging discrete token statistics to identify structurally informative anchors, offering complementary perspectives on dataset composition. As shown in \autoref{tab:imagewoof}, TGDD consistently achieves the best classification accuracy across nearly all configurations. 
Furthermore, we show in \autoref{fig:DatasetQualityScore} that the token-level statistics of TGDD surrogate accurately predict its validation accuracy via the structural score. 



Extending the evaluation to more challenging benchmarks, \autoref{tab:image-100} and \autoref{tab:image-1k} report distillation results on ImageNet-100 and ImageNet-1k in two IPC settings. TGDD consistently achieves top-tier accuracy across these configurations. These results highlight the scalability of TGDD to a more diverse and complex dataset with greater class variability. In addition, the method performs consistently well across both lightweight and deeper architectures, indicating strong generalization across model architectures.

\begin{table*}[t]
\centering
\caption{Comparison of dataset distillation performance on ImageNet-100. The best results are marked in \textbf{bold}.}
\label{tab:image-100}
\small
\setlength{\tabcolsep}{4pt}
\resizebox{1.0\textwidth}{!}{%
\begin{tabular}{ll|cccccc|c}
\toprule
IPC (Ratio) & Test Model & Random & Herding~\cite{Herding_welling} & IDC-1~\cite{IDC_Kim} & MiniMax~\cite{Minimax_diffusion_gu} & MGD$^3$~\cite{MGD3} & \textbf{TGDD (Ours)} & Full \\
\midrule
\multirow{3}{*}{10 (0.8\%)} &
ConvNet-6     & $17.0_{\pm0.3}$ & $17.2_{\pm0.3}$ & $24.3_{\pm0.5}$ & $22.3_{\pm0.5}$ & $23.4_{\pm0.9}$ & $\bm{24.8}_{\pm0.4}$ & $79.9_{\pm0.4}$ \\
& ResNetAP-10   & $19.1_{\pm0.4}$ & $19.8_{\pm0.3}$ & $25.7_{\pm0.1}$ & $24.8_{\pm0.2}$ & $25.8_{\pm0.5}$ & $\bm{27.0}_{\pm0.4}$ & $80.3_{\pm0.2}$ \\
& ResNet-18     & $17.5_{\pm0.5}$ & $16.1_{\pm0.2}$ & $25.1_{\pm0.2}$ & $22.5_{\pm0.3}$ & $23.6_{\pm0.4}$ & $\bm{24.7}_{\pm0.9}$ & $81.8_{\pm0.7}$ \\
\midrule
\multirow{3}{*}{20 (1.6\%)} &
ConvNet-6     & $24.8_{\pm0.2}$ & $24.3_{\pm0.4}$ & $28.8_{\pm0.3}$ & $29.3_{\pm0.4}$ & $30.6_{\pm0.4}$ & $\bm{31.8}_{\pm0.5}$ & $79.9_{\pm0.4}$ \\
& ResNetAP-10   & $26.7_{\pm0.5}$ & $27.6_{\pm0.1}$ & $29.9_{\pm0.2}$ & $32.3_{\pm0.1}$ & $33.9_{\pm1.1}$ & $\bm{35.2}_{\pm0.3}$ & $80.3_{\pm0.2}$ \\
& ResNet-18     & $25.5_{\pm0.3}$ & $24.7_{\pm0.1}$ & $30.2_{\pm0.2}$ & $31.2_{\pm0.1}$ & $32.6_{\pm0.4}$ & $\bm{33.4}_{\pm0.3}$ & $81.8_{\pm0.7}$ \\
\bottomrule
\end{tabular}
}
\end{table*}

\begin{table}[t]
\centering
\caption{Comparison of dataset distillation performance on ImageNet-1k. The best results are marked in \textbf{bold}.}
\label{tab:image-1k}
\small
\setlength{\tabcolsep}{5pt}
\resizebox{0.6\textwidth}{!}{
\begin{tabular}{c|cccccc}
\toprule 
IPC& SRe$^2$L~\cite{SReL_Yin} & RDED~\cite{RDED_Sun} & MiniMax~\cite{Minimax_diffusion_gu} & MGD$^3$~\cite{MGD3} &\textbf{TGDD} \\
\midrule
10 & $21.3_{\pm0.6}$ & $42.0_{\pm0.1}$ & $44.3_{\pm0.5}$ & $45.6_{\pm0.1}$ &$\bm{45.8}_{\pm0.2}$\\

50 & $46.8_{\pm0.2}$ & $56.5_{\pm0.1}$ & $58.6_{\pm0.3}$ & $60.2_{\pm0.1}$ & $\bm{60.3}_{\pm0.1}$\\
\bottomrule
\end{tabular}
}
\end{table}



The effectiveness of TGDD is further evaluated on the ImageNette and ImageIDC subsets across varying IPC settings using a ResNet-10 backbone. These datasets consist of coarser-grained, well-separated classes with lower inter-class ambiguity compared to fine-grained datasets like ImageWoof.
Experimental results in \autoref{tab:imageIDCNette} show that TGDD achieves the highest accuracy across all configurations.
In low-IPC settings, limited data makes it harder to recover semantic diversity. TGDD maintains a clear advantage by selectively choosing structurally sufficient anchors instead of using all cluster members. This reduces noise and leads to cleaner and more representative cluster centers in the low IPC setting. In contrast, methods such as MGD$^3$ \cite{MGD3}, which include all members of the cluster when computing centers, are more likely to introduce noisy samples, especially under tight data budgets. 
Beyond quantitative performance, we further provide qualitative comparisons and efficiency analysis in Appendix \S\ref{sec:Supp-AnalysisTGDD}. These additional evaluations confirm that TGDD maintains competitiveness in both sample quality and computational overhead.

\begin{table}[t]
\centering
\caption{Performance on the ImageNet subsets (Nette and IDC) across multiple images-per-class (IPC) settings. All results are obtained on a ResNetAP-10. The best results are marked in \textbf{bold}.}
\label{tab:imageIDCNette}
\small
\setlength{\tabcolsep}{5pt}
\resizebox{0.8\linewidth}{!}{
\begin{tabular}{ll|ccccc|c}
\toprule
& IPC & Random & DiT \cite{DiT_Peebles} & MiniMax \cite{Minimax_diffusion_gu} & MGD$^3$ \cite{MGD3} & \textbf{TGDD} & Full \\
\midrule
\multirow{3}{*}{Nette} & 10           & $54.2_{\pm1.6}$    & $59.1_{\pm0.7}$         &  $62.0_{\pm0.2}$        & $66.4_{\pm2.4}$ & $\mathbf{67.8}_{\pm0.6}$ &\multirow{3}{*}{$93.3_{\pm0.1}$}         \\
& 20           & $63.5_{\pm0.5}$    & $64.8_{\pm1.2}$         & $66.8_{\pm0.4}$        & $71.2_{\pm0.5}$          & $\mathbf{73.6}_{\pm0.5}$ \\
& 50           & $76.1_{\pm1.1}$    & $73.3_{\pm0.9}$         & $76.6_{\pm0.2}$        & $79.5_{\pm1.3}$          & $\mathbf{81.3}_{\pm0.6}$  \\
\midrule
\multirow{3}{*}{IDC}   & 10           & $48.1_{\pm0.8}$    & $54.1_{\pm0.4}$         & $53.1_{\pm0.2}$        & $55.9_{\pm2.1}$          & $\mathbf{57.1}_{\pm1.6}$ &\multirow{3}{*}{$92.1_{\pm0.4}$} \\
               & 20           & $52.5_{\pm0.9}$    & $58.9_{\pm0.2}$         & $59.0_{\pm0.4}$        & $61.9_{\pm0.9}$          & $\mathbf{63.4}_{\pm0.5}$ \\
               & 50           & $68.1_{\pm0.7}$    & $64.3_{\pm0.6}$         & $69.6_{\pm0.2}$        & $72.1_{\pm0.8}$          & $\mathbf{73.1}_{\pm0.8}$\\
\bottomrule
\end{tabular}
}
\end{table}

\subsection{Ablation Study}\label{sec:ablation}

\begin{table}[t]
\centering
\caption{The ablation study of the proposed token-guided dataset distillation scheme. Results are reported on the ImageWoof dataset with 20 and 50 images per class.}
\label{tab:comp-module}
\setlength{\tabcolsep}{5pt}
\resizebox{0.6\textwidth}{!}{%
\begin{tabular}{ccc|cc}
\toprule Discrete Space & PCA & Anchor Selection & IPC=20 & IPC=50 \\
\midrule
-- & -- & -- & $43.6_{\pm1.6}$ & $56.5_{\pm1.9}$ \\
\checkmark & -- & -- & $44.3_{\pm1.2}$ &  $57.4_{\pm1.3}$\\ 
\checkmark & \checkmark & -- & $45.2_{\pm1.3}$ & $58.9_{\pm1.0}$  \\
\checkmark & \checkmark & \checkmark & $\mathbf{46.3}_{\pm0.8}$ & $\mathbf{60.3}_{\pm0.9}$  \\

\bottomrule
\end{tabular}
}
\end{table}

\paragraph{Contributions of TGDD components}
\autoref{tab:comp-module} presents an ablation study evaluating the contributions of three key components in the proposed guided diffusion pipeline: discrete space, PCA, and anchor selection. Each module contributes to a distinct aspect of the framework's overall effectiveness. Clustering in the discrete space provides a symbolic representation of image content, which facilitates guidance with more specific visual concepts. PCA reduces the dimensionality of token features, making subsequent computations more robust and highlighting the most informative axes of variation. Anchor selection further enhances performance by filtering out noisy or redundant samples and retaining only structurally sufficient examples to guide synthesis. To verify that this gain comes from the structural score itself, we further evaluate two control settings on ImageWoof IPC=20. Replacing score-based selection with random-$M$ sampling within each discrete cluster drops accuracy to 43.1\%, even below the centroid baseline of 45.2\% in \autoref{tab:comp-module}. Conversely, applying score-based ranking on continuous-feature clustering yields 44.5\%, which improves over MGD$^3$ at 43.6\% but remains below the full discrete pipeline at 46.3\%. These results confirm that each component contributes incremental gains, that the score identifies meaningfully better anchors, and that discrete clustering and score-based selection are complementary.

\begin{table}[t]
\centering
\caption{Ablation study of structural score components in TGDD. Top-1 test accuracies (\%) on ImageWoof and ImageNette under IPC = 10 and 50 are reported.}
\label{tab:anchor}
\small
\setlength{\tabcolsep}{6pt}
\resizebox{0.7\textwidth}{!}{%
\begin{tabular}{ccc|cccc}
\toprule 
\multirow{2}{*}{JSD} & \multirow{2}{*}{HHI} & \multirow{2}{*}{COV}
  & \multicolumn{2}{c}{ImageWoof} & \multicolumn{2}{c}{ImageNette} \\
  &  &  & IPC=10 & IPC=50 & IPC=10 & IPC=50\\

\midrule
\checkmark & -- & -- & $38.2_{\pm2.4}$ &$58.7_{\pm0.6}$ & $65.1_{\pm1.5}$ & $80.3_{\pm0.6}$ \\
-- & \checkmark & -- & $37.8_{\pm2.3}$ &$58.7_{\pm1.0}$ & $65.9_{\pm1.0}$ & $80.5_{\pm0.8}$\\
-- & -- & \checkmark & $37.1_{\pm2.6}$ &$57.8_{\pm2.1}$ & $65.6_{\pm1.7}$ & $80.0_{\pm1.2}$\\
\checkmark & \checkmark & -- & $38.7_{\pm3.4}$ &$59.3_{\pm0.5}$ & $65.9_{\pm0.6}$ & $81.2_{\pm0.5}$ \\
\checkmark & -- & \checkmark & $38.9_{\pm1.4}$ &$59.0_{\pm1.1}$ & $65.8_{\pm0.3}$ & $80.7_{\pm1.1}$ \\
-- & \checkmark & \checkmark & $38.9_{\pm1.7}$ &$59.5_{\pm0.6}$ & $66.0_{\pm0.7}$ & $80.8_{\pm0.5}$ \\

\checkmark & \checkmark & \checkmark & $\mathbf{41.2}_{\pm2.6}$ &$\mathbf{60.3}_{\pm0.9}$ & $\mathbf{67.8}_{\pm0.6}$ & $\mathbf{81.3}_{\pm0.6}$ \\

\bottomrule
\end{tabular}
}
\end{table}

\paragraph{Structural score component}
To analyze how each structural score component in the anchor selection module contributes to dataset distillation, an ablation study is conducted, with results summarized in \autoref{tab:anchor}. The results indicate that relying on any single metric provides limited benefit, as each captures only a partial aspect of the structural score. JSD measures distributional divergence within each class, but offers limited benefit when intra-class similarity is high. COV captures class-specific compositional information, which is valuable in low-IPC settings, where preserving class identity is even more critical. HHI reflects information concentration and balance, contributing consistently across different configurations. Combining metrics in pairs leads to moderate improvements, suggesting partial complementarity. The full integration of JSD, HHI, and COV yields the best results across all the settings. These findings highlight the complementary nature of the three components and underscore the importance of their joint use for effective anchor selection in token-guided dataset distillation.

\begin{table}[t]
\setlength{\tabcolsep}{4pt}
\centering
\caption{(a) The ablation study on the effectiveness of PCA. Results are reported on ImageWoof with IPC= 10 and 50. (b) Impact of anchor set size on distillation performance. The experiment is conducted on ImageNette using ResNetAP-10 under IPC = 10 and 50.}
\vspace{-16pt}
\begin{subtable}{0.49\textwidth}
\centering
\caption{}
\label{tab:pca-d}
\vspace{-8pt}
\resizebox{0.95\textwidth}{!}{%
\begin{tabular}{c|ccccc}
\toprule
\multirow{2}{*}{IPC} & \multicolumn{5}{c}{PCA Dimension} \\
& None & 1024 & 512 & 256 & 128 \\
\midrule
10 & $38.7_{\pm0.9}$ & $40.1_{\pm1.7}$ & $\mathbf{40.2}_{\pm0.9}$ & $39.5_{\pm1.1}$ & $38.5_{\pm0.9}$ \\
50 & $57.4_{\pm1.3}$ & $57.5_{\pm1.7}$ & $\mathbf{58.9}_{\pm1.0}$ & $58.7_{\pm0.4}$ & $58.5_{\pm1.9}$ \\

\bottomrule
\end{tabular}
}
\end{subtable}
\hfill
\begin{subtable}{0.49\textwidth}
\centering
\caption{}
\label{tab:anchor_number}
\vspace{-8pt}
\resizebox{0.95\textwidth}{!}{%
\begin{tabular}{c|ccccc}
\toprule 
\multirow{2}{*}{IPC} & \multicolumn{5}{c}{Number of anchors} \\
& 1 & 5 & 10 & 20 & 30\\
\midrule
10 & $61.3_{\pm2.4}$ & $63.8_{\pm1.8}$ & $64.2_{\pm0.9}$ & $\mathbf{67.8}_{\pm0.6}$ & $64.0_{\pm1.4}$ \\
50 & $78.5_{\pm1.0}$ & $79.7_{\pm0.6}$ & $\mathbf{81.3}_{\pm0.6}$ & $79.9_{\pm0.1}$ & $80.2_{\pm1.0}$ \\
\bottomrule
\end{tabular}
}
\end{subtable}
\end{table}

\paragraph{Dimensions of PCA}
\autoref{tab:pca-d} investigates the impact of varying the output dimensionality of PCA within the proposed framework. Results are reported on ImageWoof, with ResNetAP-10 trained at IPC 10 and 50. The baseline model without PCA achieves 57.4\% accuracy. Applying PCA leads to consistent improvements, with the best performance observed at 512 dimensions. However, further compressing the feature dimensionality to 256-D and 128-D results in a slight performance decline, suggesting that excessive compression can remove important semantic information. These results confirm the effectiveness of PCA as a feature compression mechanism and indicate that a 512-dimensional representation offers the best trade-off between compactness and task performance.

\paragraph{Effect of anchor quantity on generation}
To identify the optimal number of anchors for guiding the distillation process, we study how varying the anchor set size affects generation quality. Results are shown in \autoref{tab:anchor_number}. When IPC = 10, using 20 anchors yields the best performance, while at higher IPC levels, fewer anchors (\eg, 10) are sufficient. This trend aligns with expectations: under limited data conditions, using more guiding anchors helps enrich the semantic coverage of the generated samples. However, when excessive anchors are used, they may introduce noise or redundancy, leading to a drop in performance.




\begin{table}[t]
\centering
\caption{Performance comparison across various representation spaces and diffusion architectures. Using our structural score for anchor selection, discrete tokenizers consistently outperform continuous models in guiding the generation process on ImageWoof.}
\label{tab:cross_archi}
\small
\setlength{\tabcolsep}{6pt}
\resizebox{1\textwidth}{!}{
\begin{tabular}{@{}c|c|ccc|ccc@{}}
\toprule
\multirow{2}{*}{Diffusion Arch.} & \multirow{2}{*}{No Guidance} & \multicolumn{3}{c|}{Continuous Models} & \multicolumn{3}{c}{Discrete Tokenizers} \\
 &  & VAE \cite{LDM_Rombach} & CLIP \cite{CLIP_radford} & DINOv2 \cite{DINOV2_Oquab} & VQGAN \cite{VQGAN_Esser} & BEiTv2 \cite{BeitV2_Peng} & VQ-VAE \cite{VAR_Tian} \\
 \midrule
DiT \cite{DiT_Peebles} & $49.3_{\pm0.2}$ & $56.5_{\pm1.9}$ & $57.5_{\pm0.4}$ & $58.1_{\pm0.6}$ & $60.0_{\pm0.4}$ & $59.3_{\pm0.7}$ & $60.3_{\pm0.9}$ \\
LDM \cite{LDM_Rombach} & $48.6_{\pm1.6}$ & $52.9_{\pm2.2}$ & $53.4_{\pm1.3}$ & $53.7_{\pm0.8}$ & $55.8_{\pm0.7}$ & $54.2_{\pm0.5}$ & $55.1_{\pm0.6}$ \\
\bottomrule
\end{tabular}
}
\vspace{-12pt}
\end{table}

\paragraph{Generalization across spaces and diffusion architectures}
To validate the generalization and effectiveness of the proposed TGDD framework, we conduct a comprehensive cross-architecture evaluation. We evaluate our method using three distinct discrete visual tokenizers: VQ-VAE \cite{VAR_Tian}, VQGAN \cite{VQGAN_Esser}, and BEiTv2 \cite{BeitV2_Peng}, and compare them with widely used continuous embedding models including VAE encoder\cite{VAE_original} from \cite{LDM_Rombach}, CLIP \cite{CLIP_radford}, and DINOv2 \cite{DINOV2_Oquab}. These representations are paired with two diffusion architectures, specifically the transformer-based DiT \cite{DiT_Peebles} and the U-Net-based LDM \cite{LDM_Rombach}. For continuous models, we follow the standard practice of clustering dense features and selecting samples closest to the centroids as anchors \cite{MGD3}. The results in \autoref{tab:cross_archi} demonstrate the robustness of our token-guided approach. While applying guidance with advanced continuous models like DINOv2 and CLIP improves upon the unguided baselines, they are consistently outperformed by our TGDD framework across all tested discrete tokenizers and generative architectures. This empirical evidence confirms that our structural score metrics actively drive performance gain and provide a robust anchor selection strategy that generalizes across diverse discrete tokenizers and offers a novel compositional perspective to the dataset distillation task.





\section{Conclusion}
\label{sec:Conclusion}

In this work, we revisit dataset distillation through the compositional perspective of discrete visual tokens. By mapping image representations to discrete tokens via visual tokenizers, we reframe the assessment of distilled datasets as a matter of structural score rather than merely distributional similarity. We find that the statistics of visual tokens can provide a reliable predictor of validation performances. 
Building on this insight, we further leverage token statistics to guide diffusion denoising. 
The proposed Token-Guided Dataset Distillation (TGDD) achieves state-of-the-art performance across multiple benchmarks.
These results suggest that discrete token analysis provides principled value for understanding and guiding dataset distillation.


\section*{Acknowledgements}
This work was supported by Mitacs through the Mitacs Accelerate Program (Grant No. IT42711), by TerraSense Analytics, and in part by the U.S. National Science Foundation (OAC-2118240, HDR Institute: Imageomics).

%
%
\bibliographystyle{splncs04}
\bibliography{main}

\clearpage

\appendix
\renewcommand{\theHsection}{A\arabic{section}}



\section*{Appendix}

The appendix is organized as follows:
\begin{itemize}
    \item \S\ref{sec:supp-algorithm}: Pseudo-code for the anchor selection procedure used in TGDD.
    \item \S\ref{sec:supp-ImplementationDetails}: Further implementation details for our experimental setup.
    \item \S\ref{sec:Supp-AnalysisTGDD}: Additional analysis of the proposed TGDD model, including different scale weighting configurations, efficiency analysis, and qualitative comparisons.
\end{itemize}

\section{Anchor Selection Algorithm}
\label{sec:supp-algorithm}

The anchor selection procedure is summarized in \autoref{alg:anchor_selection}. The algorithm takes the tokenized dataset as input and selects the images with the top structural scores to guide the diffusion generation process. The output consists of the top $M$ anchors for each cluster, where $M = 10$ for IPC settings greater than 50 and $M = 20$ for IPC settings below 50.

\begin{algorithm}[h]
\caption{Token-Guided Anchor Selection}
\label{alg:anchor_selection}
\footnotesize
\SetAlgoVlined
\DontPrintSemicolon
\setlength{\algomargin}{0pt}
\SetInd{0.3em}{0.6em}
\SetKwInOut{Input}{Input}\SetKwInOut{Output}{Output}
\Input{Dataset $\mathcal{T}=\{(x_i,y_i)\}$; multi-scale VQ-VAE; codebook size $V$; scale weights $\{w_\ell\}_{\ell=1}^{L}$; IPC $\Rightarrow$ clusters $K_c$ per class; \textbf{feature weights $W_{\mathrm{JSD}},W_{\mathrm{HHI}},W_{\mathrm{Cov}}$}; TF-IDF top-token size $K_{\text{tfidf}}$; PCA dim $d$; Anchor number $M$.}
\Output{Anchor set $\mathcal{A}$ with $M$ anchors per cluster.}

\For{each class $c=1,\dots,C$}{
  \For{each $x_i$ with $y_i=c$}{
    Compute per-scale token histograms $\vp_i^{(\ell)}\!\in\!\mathbb{R}^{V}$; 
    fuse $\vp_i \leftarrow \sum_{\ell=1}^{L} w_\ell\,\vp_i^{(\ell)}$\;
  }
  Compute per-scale TF, classwise IDF; form fused TF--IDF vector ; set $\mathcal{T}_c$ of size $K_{\text{tfidf}}$\;
  Apply PCA to $d$ dims, L2-normalize; run $k$-means into $K_c$ clusters to obtain member sets $\{\mathcal{S}_{c,m}\}_{m=1}^{K_c}$.\;
  \For{each $m=1,\dots,K_c$}{
    Compute the cluster prior $\boldsymbol{\mu}_{c,m}$ from members $\mathcal{S}_{c,m}$\;
    For $i\!\in\!\mathcal{S}_{c,m}$, set $\mathbf{m}{=}\tfrac{1}{2}(\vp_i{+}\boldsymbol{\mu}_{c,m})$ and compute
    $\mathrm{JS}(\vp_i,\boldsymbol{\mu}_{c,m}){=}\tfrac{1}{2}\big(\mathrm{KL}(\vp_i\|\mathbf{m}){+}\mathrm{KL}(\boldsymbol{\mu}_{c,m}\|\mathbf{m})\big)$\;
    For $i\!\in\!\mathcal{S}_{c,m}$, compute $\mathrm{Cov}_i=\sum_{k\in\mathcal{T}_c} p_i(k)$ and $D_i=1-\sum_{k=1}^{V}\!\big(p_i(k)\big)^2$; 
    \textbf{normalize} $\{\mathrm{JS}_i\}$, $\{\mathrm{Cov}_i\}$ and $\{D_i\}$ within each cluster\;
    Rank by $s_i= w_{\mathrm{Cov}}\,\widetilde{\mathrm{Cov}}_i + w_{\mathrm{HHI}}\,\widetilde{D}_i + w_{\mathrm{JSD}}\,(1-\widetilde{\mathrm{JS}}_i)$ and add the \textbf{top $M$} indices to $\mathcal{A}$ for cluster $m$\;
  }
}
\end{algorithm}

\section{Additional Implementation Details}
\label{sec:supp-ImplementationDetails}

Throughout this study, we follow the literature \cite{Minimax_diffusion_gu,MGD3} and evaluate the effectiveness of the generated datasets using three main architectures: \textbf{ConvNet-6}, a 6-layer convolutional network with 256$\times$256 input size; \textbf{ResNetAP-10}, a 10-layer ResNet \cite{resnet_he} with average pooling for downsampling; and \textbf{ResNet-18}, an 18-layer ResNet with instance normalization. For experiments on ImageWoof, ImageNette, ImageIDC, and ImageNet-100, we adopt the hard-label protocol used in \cite{MGD3}. All architectures are trained using SGD with a learning rate of 0.01 and a decay factor of 0.2 applied at 2/3 and 5/6 of the total training epochs. Models are trained for 1500 epochs under IPC values of 20, 50, 70, and 100, and for 2000 epochs under the IPC=10 setting. Random resize-crop and CutMix augmentations are applied during training.

For evaluation on ImageNet-1k, we use the soft-label protocol described in \cite{Minimax_diffusion_gu,MGD3}. Training is performed on a ResNet-18 architecture serving as both the teacher and the student for 300 epochs. The AdamW optimizer is used for ImageNet-1K training, with a learning rate of 0.001 and a weight decay of 0.01.

\section{Additional Analysis of TGDD}
\label{sec:Supp-AnalysisTGDD}

\subsection{Different Scale Weighting Configurations}
\label{sec:supp-ScaleWeightingConfig}

Our TGDD framework employs a multi-scale VQ-VAE \cite{VAR_Tian} as the default visual tokenizer. Due to its pyramidal architecture, the token count grows quadratically at each subsequent scale. Therefore, a direct summation of the distributions causes finer scales to dominate the result. To mitigate this inherent density imbalance, we performed an ablation study on the ImageWoof and ImageNette datasets by grouping the 10 scales into Low, Mid, and High bands. As shown in \autoref{tab:VQVAE-weight}, we observe a consistent trend: configurations that prioritize coarse scales (Low > Mid > High) outperform uniform or inverse weighting schemes. Based on these empirical results, we adopted [3, 1, 0.5] as the default configuration to maintain an optimal balance between the information from different scales.


\begin{table}[]
\centering
\caption{The results demonstrate the sensitivity of the model to the relative importance assigned to Low, Mid, and High resolution scales.}
\label{tab:VQVAE-weight}
\small

\begin{tabular}{c|ccccc}
\toprule 
\multirow{2}{*}{Benchmark} & \multicolumn{5}{c}{Scale Weighting Configurations} \\

& [5, 1, 0.1] & [3, 1, 0.5] & [1, 1, 1] & [1, 2, 3] & [0.5, 1, 3] \\
\midrule
ImageWoof & $59.1_{\pm0.3}$ & $\bm{60.3}_{\pm0.9}$ & $58.8_{\pm1.4}$ & $58.9_{\pm0.4}$ &$58.2_{\pm1.7}$\\

ImageNette & $80.5_{\pm0.6}$ & $\bm{81.3}_{\pm0.6}$ & $80.1_{\pm0.6}$ & $79.2_{\pm0.7}$ & $79.2_{\pm0.9}$\\
\bottomrule
\end{tabular}

\end{table}

\subsection{Efficiency Analysis}
\label{sec:supp-Efficiency}
We analyze the distillation time of TGDD by decomposing the computation into tokenization, PCA transformation, anchor selection, and diffusion-based generation. TGDD uses the same diffusion backbone as existing methods (e.g., Minimax Diffusion \cite{Minimax_diffusion_gu} and MGD$^3$ \cite{MGD3}) and operates without any fine-tuning, so the main additional computational cost comes from the pre-processing stages. The total complexity can be written as
\[
\mathcal{C}_{\text{TGDD}}
= \mathcal{C}_{\text{Diff}}
+ \mathcal{O}(N T)
+ \mathcal{O}(d^{2} N)
+ \mathcal{O}(N d)
+ \mathcal{O}(N \log K),
\]
where each term corresponds to a specific component of TGDD. The diffusion sampling cost $\mathcal{C}_{\text{Diff}}$ is shared across diffusion-based distillation methods and dominates the overall runtime. The tokenization step processes $N$ images, each represented by $T$ discrete visual tokens, leading to the $\mathcal{O}(N T)$ term. PCA projection is applied to all token histograms, and computing a $d$-dimensional projection requires multiplying by a $d\times d$ matrix, resulting in $\mathcal{O}(d^{2}N)$. Anchor selection evaluates structural scores, including JSD, HHI, and COV, each computed over a $d$-dimensional feature, producing the $\mathcal{O}(N d)$ term, followed by a partial ranking step that contributes $\mathcal{O}(N \log K)$. Here, $K$ denotes the number of anchors retained per class for guiding the diffusion process. These additional operations are lightweight compared with $\mathcal{C}_{\text{Diff}}$.

To empirically validate the efficiency, \autoref{tab:EfficiencyAnalysis} presents the per-image processing time for each stage of the TGDD and MGD$^3$ pipelines.
As demonstrated in the table, the computational cost of feature extraction and anchor selection is minor, with the diffusion-based generation process accounting for the majority of the total runtime. This confirms that the generation phase is the primary computational bottleneck, whereas the proposed token-based assessment introduces limited overhead. Specifically, for the total distillation of the 10-class ImageWoof dataset on a single RTX 3080 GPU, TGDD requires approximately 0.43 hours. This duration is comparable to the highly efficient MGD$^3$ at 0.32 hours, and is significantly faster than other diffusion-based baselines that require training, such as Minimax Diffusion which demands 2.02 hours. Considering the substantial performance gain of up to 5.4\% on ImageWoof, TGDD delivers stronger performance with only modest extra preprocessing effort.

\begin{table}[h]
\centering
\caption{Comparison of per-image processing time between TGDD and MGD$^3$}
\label{tab:EfficiencyAnalysis}
\small
\resizebox{0.5\textwidth}{!}{
\begin{tabular}{c|ccc}
\toprule 
Model & Feature Extraction & Clustering & Generation \\
\midrule
TGDD & 0.0198s & 0.0344s & 1.7s \\

MGD$^3$ & 0.0212s & 0.0010s & 1.7s\\
\bottomrule
\end{tabular}
}
\end{table}

\subsection{Visualization of Synthetic Samples}\label{sec:supp-Qualitative_Comparison}
\begin{figure*}[h]
  \centering

  \begin{subfigure}{\linewidth}
    \centering
    \includegraphics[width=\linewidth]{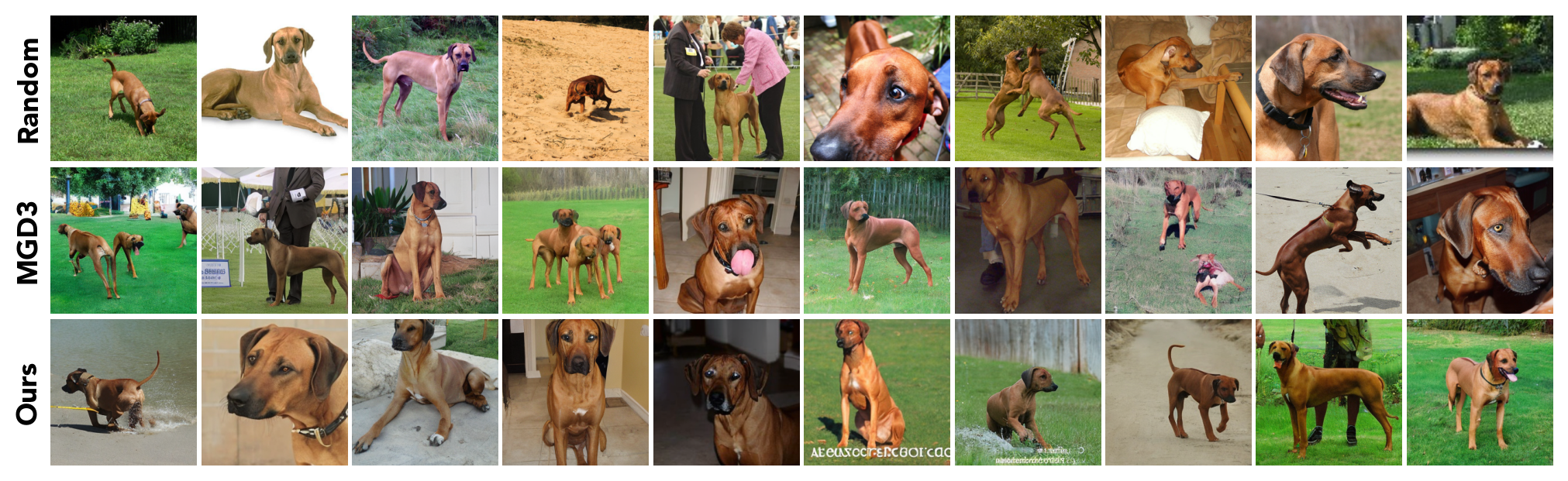}
    \caption{Rhodesian Ridgeback}
  \end{subfigure}

  \begin{subfigure}{\linewidth}
    \centering
    \includegraphics[width=\linewidth]{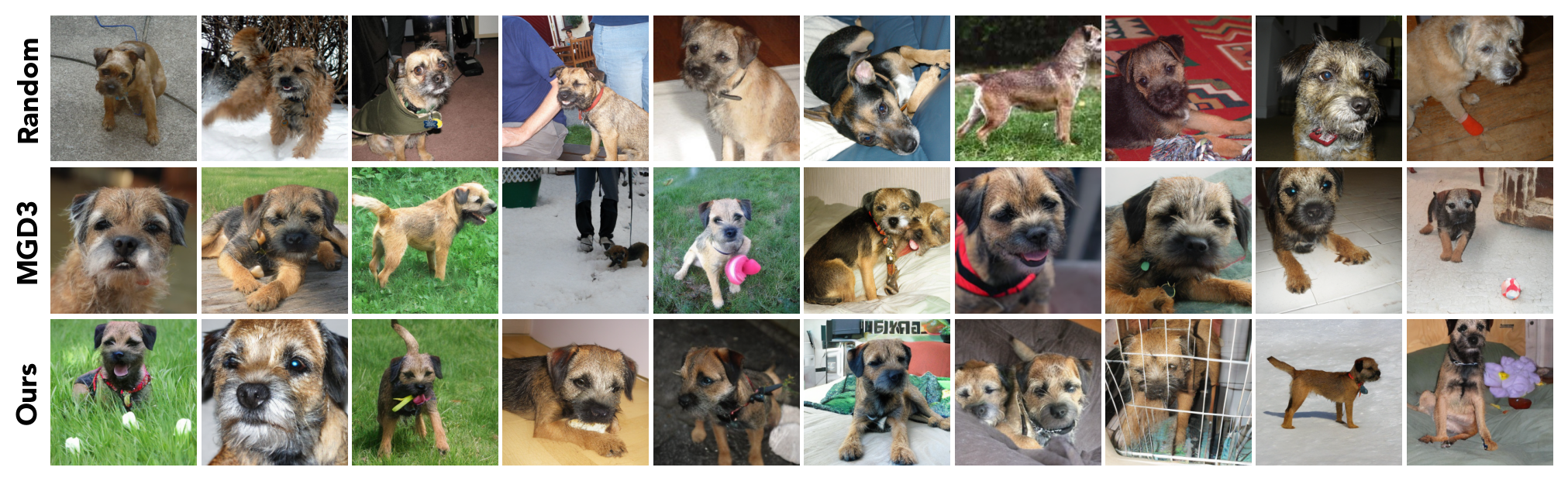}
    \caption{Border Terrier}
  \end{subfigure}

  \begin{subfigure}{\linewidth}
    \centering
    \includegraphics[width=\linewidth]{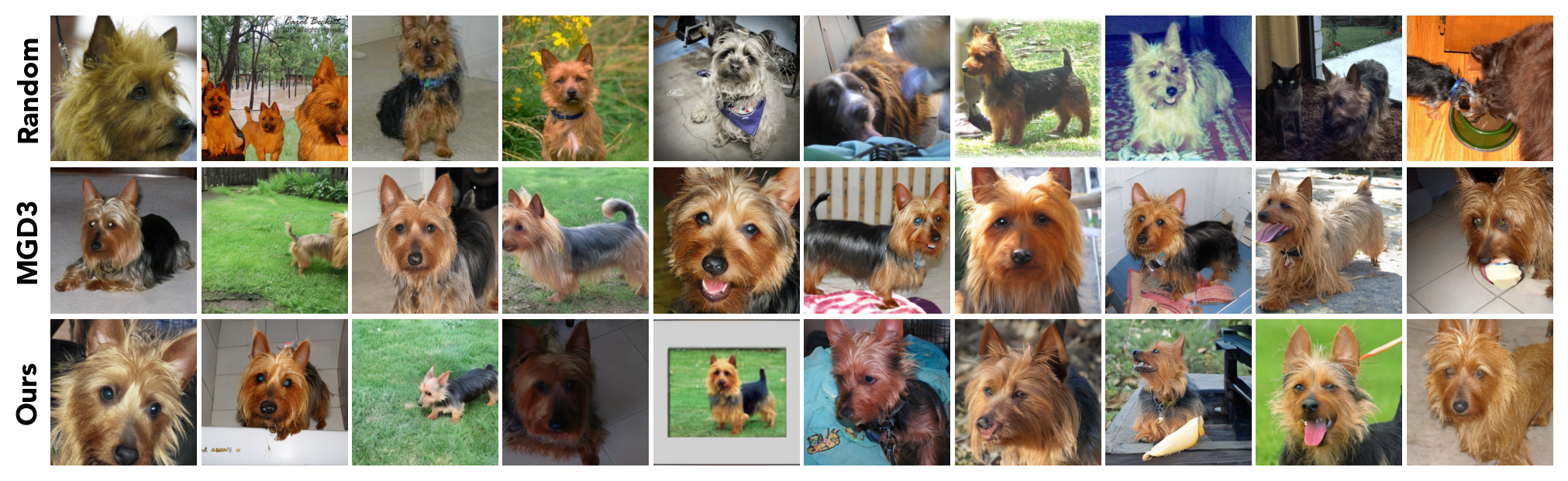}
    \caption{Australian Terrier}
  \end{subfigure}

  \caption{Visualization of generated samples for three ImageWoof classes (Rhodesian Ridgeback, Border Terrier, and Australian Terrier). For each class, we show real images, images generated by MGD$^3$ \cite{MGD3}, and images generated by TGDD under the IPC = 10 setting with the same random seed.}
  \label{fig:qualitative_three}
\end{figure*}

We visualized the images generated by TGDD on the ImageWoof dataset alongside those produced by MGD$^3$ \cite{MGD3} and real samples randomly selected from the original dataset, as shown in \autoref{fig:qualitative_three}. All results are obtained under the IPC = 10 setting using the same random seed. The proposed TGDD generates high-quality samples that closely resemble the real images, demonstrating its ability to preserve both semantic content and structural details. While MGD$^3$ also produces high-quality synthetic data, some images exhibit artifacts, such as slight distortions in facial regions or incomplete object parts. These differences likely arise from the anchor selection strategy of TGDD, which selects only the top-ranked and structurally sufficient anchors to guide the generation. This selective guidance reduces the noise introduced by averaging large numbers of candidate images and mitigates the influence of repeated patterns or outliers that may negatively affect sample quality.

\end{document}